# Improving Deep Learning Models for Pediatric Low-Grade Glioma Tumors Molecular Subtype Identification Using 3D Probability Distributions of Tumor Location

Khashayar Namdar[1,2,6,9,10], Matthias W. Wagner[1,2,3], Kareem Kudus[1,2,6], Cynthia Hawkins[5], Uri Tabori[4], Birgit B. Ertl-Wagner[1,2,3], Farzad Khalvati[1,2,3,6,7,8,9]

Affiliations:

1. Division of Neuroradiology, Department of Diagnostic Imaging, The Hospital for Sick Children (SickKids), Toronto, ON, Canada
2. Neurosciences & Mental Health Research Program, SickKids Research Institute, Toronto, ON, Canada
3. Department of Medical Imaging, University of Toronto, Toronto, ON, Canada
4. Department of Neurooncology, The Hospital for Sick Children, Toronto, ON, Canada
5. Department of Paediatric Laboratory Medicine, Division of Pathology, The Hospital for Sick Children, University of Toronto, Canada
6. Institute of Medical Science, University of Toronto, Toronto, ON, Canada
7. Department of Computer Science, University of Toronto, Toronto, ON, Canada
8. Department of Mechanical and Industrial Engineering, University of Toronto, Toronto, ON, Canada
9. Vector Institute, Toronto, ON, Canada
10. NVIDIA Deep Learning Institute, Austin, TX, United States

**Abstract**

**Background and Purpose:** Pediatric low-grade glioma (pLGG) is the most common type of brain tumor in children, and identification of molecular markers for pLGG is crucial for successful treatment planning. Convolutional Neural Network (CNN) models for pLGG subtype identification rely on tumor segmentation. We hypothesize tumor segmentations are suboptimal and thus, we propose to augment the CNN models using tumor location probability in MRI data.

**Materials and Methods:** Our REB-approved retrospective study included MRI Fluid-Attenuated Inversion Recovery (FLAIR) sequences of 143 BRAF fused and 71 BRAF V600E mutated tumors. Tumor segmentations (regions of interest (ROIs)) were provided by a pediatric neuroradiology fellow and verified by a senior pediatric neuroradiologist.

In each experiment, we randomly split the data into development and test with an 80/20 ratio. We combined the 3D binary ROI masks for each class in the development dataset to derive the probability density functions (PDF) of tumor location, and developed three pipelines: location-based, CNN-based, and hybrid.

**Results:** We repeated the experiment with different model initializations and data splits 100 times and calculated the Area Under Receiver Operating Characteristic Curve (AUC). The location-based classifier achieved an AUC of 77.90, 95% confidence interval (CI) (76.76, 79.03). CNN-based classifiers achieved AUC of 86.11, CI (84.96, 87.25), while the tumor-location-guided CNNs outperformed the formers with an average AUC of 88.64 CI (87.57, 89.72), which was statistically significant (Student's t-test p-value 0.0018).

**Conclusion:** We achieved statistically significant improvements by incorporating tumor location into the CNN models. Our results suggest that manually segmented ROIs may not be optimal.

# Introduction

Brain tumors are the most common solid cancer among children, with pediatric Low-Grade Glioma (pLGG) being the most frequent [1][2][3]. The advent of targeted therapies such as B-Raf proto-oncogene, serine/threonine kinase (BRAF) inhibitors [4][5] has improved therapeutic outcomes of pLGG, but successful treatment planning for pLGG is governed by identifying tumor type and molecular subtype [6][7]. A precise molecular subtype identification and targeted therapy lead to five-year survival rates over 75% [8][9]. Currently, the standard of care for molecular subtype identification of pLGG is biopsy, which is invasive with potential associated complication risks and sometimes is not feasible due to a tumor's location [10][11][12][13].

While MRI could represent a non-invasive alternative to biopsy for tumor classification and depicts the tumor in its entirety, determining the molecular subtype of a tumor based on MRI remains challenging [14]. The feasibility of Machine Learning (ML) algorithms to identify molecular markers of pLGG tumors has been demonstrated, but there remain important gaps warranting further improvement, including a location-based approach [15][16].

We therefore aimed to establish and assess a tumor-location- and Convolutional Neural Network (CNN) based pipeline to identify pLGG molecular subtypes using MRI and to subsequently merge the pipelines and form an outperforming tumor-location-guided CNN algorithm. Our research presents notable advancements to the prevailing literature on imaging-based pLGG subtype determination. Specifically, our contributions encompass:

- Utilizing a repetitive approach, which is compatible with the open-radiomics protocol, to measure randomness of CNN pipelines for pLGG molecular biomarker identification.

- Using tumor location as an independent modality, as opposed to the conventional approach of using tumor location as a categorical variable.

- Enhancing CNN pipelines through the utilization of tumor location

# Related work

Previous studies on the pLGG classification using ML are limited. Nonetheless, multiple impactful clinical studies have been conducted. Bag et al. provided comprehensive details about the radiohistogenomics of pLGG, and how different factors contribute to risk stratification of the patients [17]. They identified five genetic profile, location, age at presentation, and histology as the decisive features for pLGG phenotype risk assessment. They also listed BRAF fusion and BRAF V600E point mutation (BRAF mutation) among the common gene alterations in pLGG (i.e., pLGG molecular subtype), and illustrate how the two subtypes are different in term of tumor location. In their study, while 75% of the BRAF Fusion cases were located in cerebellum, the rate for BRAF mutation was only 5%. In contrast, majority of BRAF mutation cases were observed in cerebral hemispheres (56%). As a consequence, BRAF fusion tumors are often surgically resectable, leading to excellent overall survival and low progression rates [18].

Wagner et al. used radiomics to conduct pretherapeutic differentiation of BRAF-mutated and BRAF-fused tumors [19]. In their bi-institutional retrospective study, T2 fluid-attenuated inversion recovery (FLAIR) MR images of 115 pediatric patients from 2 children's hospitals were analyzed. They extracted radiomics features from the regions of interest (ROIs) that a pediatric neuroradiologist provided, and used random forest (RF) as the binary classifier. They evaluated the models using a 4-fold cross validation on 94 patients from one institute and conducted a one-time external validation on the remaining 21 patients from the second hospital. They provided the receiver operating characteristic (ROC) curves of the models and achieved an average area under the ROC curve (AUROC) of 0.75 on the internal cohort and an AUROC of 0.85 on the external cohort. Their statistical analysis showed location (supra- versus infratentorial) and age were significant predictors of BRAF status on the first cohort. Additionally, the average AUROC was improved to 0.77 when age and location were appended to the radiomics features.

In a separate study on a larger bi-institutional dataset of 251 pLGG FLAIR MR images, Wagner et al. conducted the dataset size sensitivity analysis [20]. They showed data splits and model initialization impose randomness to the performance of the ML classifiers which impacts the results in two aspects, namely average AUROC and variance of the AUROCs. They confirmed the scale of the dataset size was reliable to train robust pipelines, but an individual global model could be biased. With only 60% of the training data, they achieved comparable results to those of models that used the entire training set, with an average AUROC of 0.83 compared to 0.85.

Haldar et al. retrospectively studied a dataset of 157 pLGG patients from the Children's Hospital of Philadelphia (CHOP), and utilized a conventional unsupervised ML algorithm to classify the subjects [21]. In unsupervised ML, the models are trained to find patterns in the data without considering the labels which makes this framework appropriate for unlabelled data. In the context of classification, it is common to use unsupervised ML for data clustering. However, data clusters and labels, in this case pLGG subtypes, may not agree, and thus supervised ML is preferred. Conforming to the clustering approach, Haldar et al. employed Principal Component Analysis (PCA) followed by K-means algorithm to assign the patient images into three subgroups. Using Kruskal-Wallis test, they demonstrated that the distribution of tumor histology and location among the three imaging clusters was different. Although PCA is sensitive to outliers and combination of the statistical analysis and unsupervised learning has a weak predictive power compared to supervised learning, the study is another evidence for the importance of location in pLGG subtype identification.

Korfiatis et al. reviewed ML approaches, including deep learning (DL), for predicting molecular markers from MRI [22]. They provided a summary of the benchmark models for different datasets of brain glioma, such as Medulloblastoma and Glioblastoma. They highlighted that DL had significantly improved prediction of markers for gliomas but lacked feature interpretability. Nevertheless, they did not provide any specific study on pLGG or tumor location.

Xu et al. analyzed a dataset of 113 pLGG patients (43 BRAF V600E mutations versus 70 cases of other subtypes) using radiomics features to identify BRAF mutations [23]. They conducted a one-time train-test splitting with the ratio of 70/30 and employed a 5-fold cross-validation to find the best setting for their pipeline using the training cohort. Multiple classifiers were evaluated, and RF outperformed all. They achieved an average training AUROC of 0.72, with 95 % confidence interval (CI), (0.602, 0.831), and a test AUROC of 0.875. Although the data splitting approach may result in irreproducible AUROCs, their conclusions are remarkable. Tumor location (supratentorial versus infratentorial) was a significant predictor of BRAF mutation and when combined with radiomics, it improved the average training AUROC to 0.754 with 95% CI of (0.645, 0.844) and the test AUROC to 0.934. Additionally, tumor location side (left versus right) was studied, and it was not a significant predictor of BRAF mutation.

## Materials and Methods

### Dataset

The local institutional review board approved this retrospective study. Due to its retrospective nature, the local research ethics board exempted the need for informed consent. The internal dataset included 143 fusion and 71 mutation cases, making the dataset one of the largest in the world. Patients were identified using the electronic health record (EHR) database of the hospital from January 2000 to December 2018. Inclusion criteria were an age of 0 to 18 years, and histopathological confirmation of BRAF status.

All patients underwent brain MRI at 1.5T or 3T, using MRI scanners from various vendors (Signa, GE Healthcare; Achieva, Philips Healthcare; Magnetom Skyra, Siemens Healthineers). We only used the axial FLAIR sequence (3-6 mm slice thickness; 3-7.5 mm gap) in order to maximize the sample size. Segmentation of ROIs was performed by a neuroradiologist using a semi-automated approach on FLAIR images with the Level-Tracing-Effect tool in the 3D Slicer library (Version 4.10.2, https://www.slicer.org/). In terms of reproducibility and robustness, the semi-automatic process has been confirmed to surpass multi-user manual delineation [24]. The final ROIs were confirmed by a pediatric neuroradiology trained and board-certified radiologist with seven years of neuroradiology research experience.

The preprocessing pipeline included labeling, resampling, normalization, skull stripping, bias correction, and registration to the SRI24 [25] atlas for each image volume. SRI24 is an MRI-based atlas based on normal adult human brain anatomy and is a well-known option for preprocessing brain MRI [26]. It should be highlighted that registration is a key step in the preprocessing pipeline, and our study relies on it.

Without proper registration, tumor location may become imprecise for pLGG molecular biomarker identification.

Location-based Analysis

In contrast to the previous that tumor location was used as a binary variable [19], we defined tumor location probability density functions (PDF) to achieve voxel-level granularity. Tumor location PDFs of BRAF V600E mutation and BRAF fusion were defined through summing and normalizing the 3D binary segmentation masks in FLAIR images for each class in the development dataset (i.e., the union of training and validation datasets). In the test cohort, the probabilities of belonging to each class were calculated by summation of a voxel-wise multiplication of the binary segmentation mask of the test case and the 3D PDF of the corresponding class in the development dataset (Eq.1). Figure 1 illustrates the projections of the PDFs in axial, coronal, and sagittal planes. We repeated the data split (80/20 for training and test) 100 times and calculate the AUROC.

$$p_c = \sum (segmentation \odot pdf_c) \mid C \in \{fusion, mutation\} \tag{1}$$

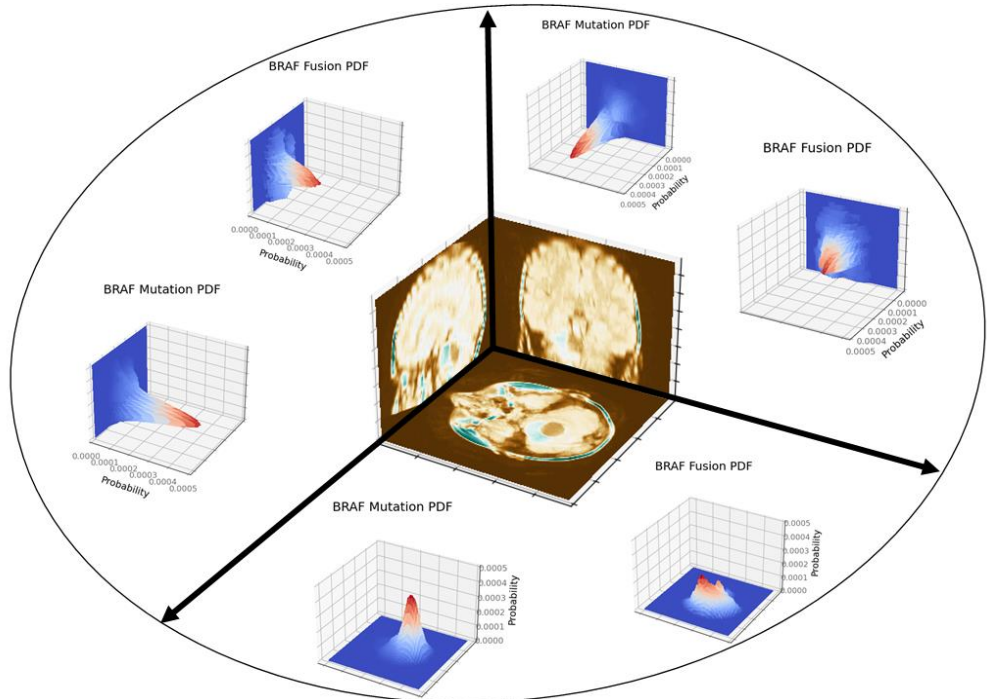

**Figure 1:** Projections of the tumor location PDFs in axial, coronal, and sagittal planes

### CNN-based Analysis

We used off-the-shelf models such as 3D ResNet [32], as well as the shallow CNN architecture described in Appendix A. To train the models, we chose a batch size of 8, maximum number of epochs of 10, learning rate of 0.1, Cross Entropy (CE) as the loss function, and stochastic gradient descent (SGD) [33] as the optimizer. The models were implemented using PyTorch 1.10.2, in a Python 3.9.7 environment with cuda 11.3. We utilized two GeForce RTX 3090 Ti GPUs on a Lambda Vector GPU workstation.

### Tumor-location-guided CNN Analysis

The setting for the tumor-location-guided CNN analysis was identical to how the CNN pipeline was developed except for the maximum number of epochs which was increased to 20 based on observations on the first five experiments. Figure 2 illustrates the three approaches we used to identify pLGG subtype.

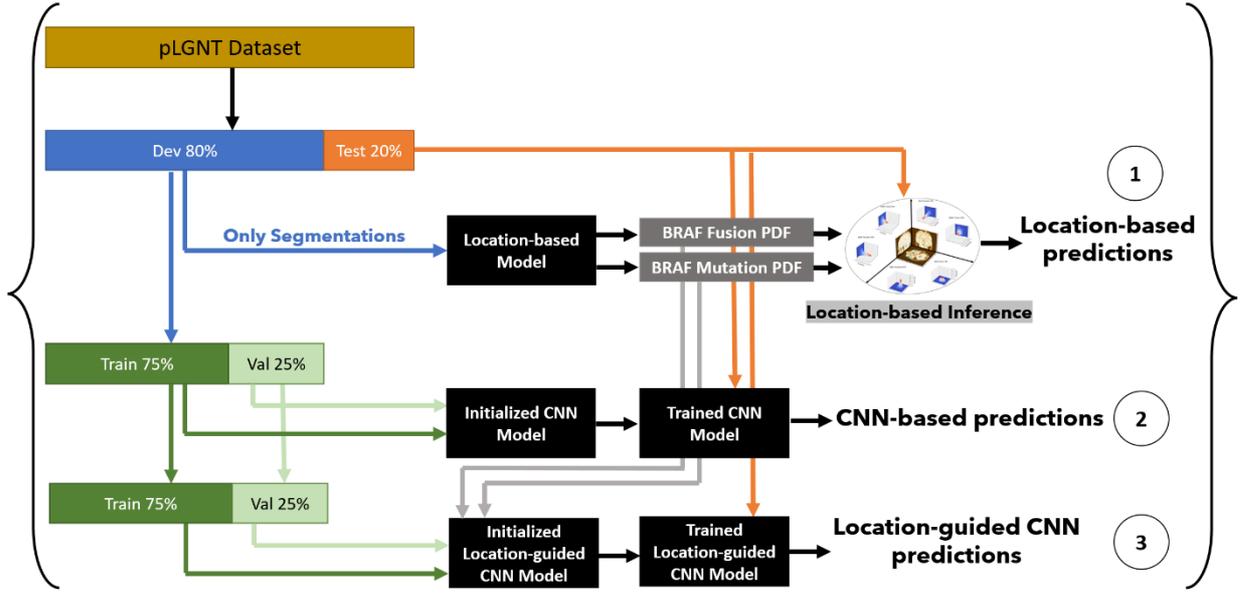

**Figure 2:** Location-based pLGNT molecular biomarker identification pipelines:
1) location-only 2) CNN-only 3) location-augmented CNN

In the tumor-location-guided CNN algorithm, the location PDFs are applied to each image according to Equation 2, where offset (0.2 in the experiments) is a scalar to help retain regions outside of the segmentation.

$$input = (offset + segmentation) \times image \times \sum_{C} p_C \times pdf_C \qquad (2)$$

$$| \; C \in \{fusion, mutation\}$$

# Results and Discussion

Radiomics and CNNs form the two established branches of ML, applicable to pLGG molecular subtype identification [27]. Radiomics refers to a set of manually defined equations through which ROIs are transformed into quantitative tabular data formats [28], [29], and ML models such as RF can be trained to identify pLGG molecular markers from radiomics features [19]. In contrast, CNNs learn to extract the features and are not limited to predefined formula for pattern recognition. Thus, CNNs have the potential to outperform radiomics-based models. In conventional CNNs, the feature extraction is done by sequential convolution layers, and fully connected (FC) layers classify the features. DL refers to utilizing CNNs with a high number of convolutional layers, and deep models are state-of-the-art (SOTA) on multiple large-scale datasets [30]. However, on the small datasets where there is no pre-trained model available, deeper models may not improve the performance [31]. In our initial experiments the shallow CNN outperformed 3D ResNet [32], and thus we used the shallow architecture throughout the research. This enabled us to lower the computational load and increase the data split repeats.

In comparison to the CNN-based analysis where ROIs were formed through element-wise multiplication of segmentations and images, two revisions are applied to design the tumor-location-guided CNN: a) the offset scalar is added to the segmentation binary mask to avoid losing the image areas where the mask elements are zero, b) the two PDFs are weighted based on their probability for a given image and summed, and the result is used as a mask to recess image areas where tumor presence is unlikely. The motivation for implementing the tumor-location-guided CNN algorithm is utilizing regions outside the manual segmentation to improve the classification performance.

Nevertheless, simpler methods could be employed, and prior to implementing the pipeline, we tried different approaches for incorporating tumor location information into the CNNs. However, we achieve marginal or no improvement in terms of average AUROC. Injecting location-based probabilities at different layers of the CNN architecture, and ensemble of the CNN and location-based models were among the methods that did not help.

There are multiple sources of variability impacting generalizability and reproducibility of ML pipelines [34]. Classifiers are sensitive to any form of change in input, and different scanner vendors. Imaging device settings, imaging protocols, contouring discrepancy (known as intra- and inter-reader variability), image normalization, and CNN initialization may result in variation in results. We employ the repetitive approach proposed in OpenRadiomics [35] for reproducible ML research on relatively small datasets. When the order of dataset size is of hundred, outliers might impact the fairness of data splits [36]. Hence, creating a single reproducible model becomes infeasible and the focus should be on training repeatable pipelines.

We repeated the experiments 100 times and unified the data splits for all the three pipelines. The location-based classifier achieved an AUROC of 77.90, 95% CI (76.76, 79.03). CNN-based classifiers landed at AUROC of 86.11, CI (84.96, 87.25), and the tumor-location-guided CNNs surpassed the other two with an average AUROC of 88.64 CI (87.57, 89.72). Figure 3 shows the results for the three pLGG subtype identification methods.

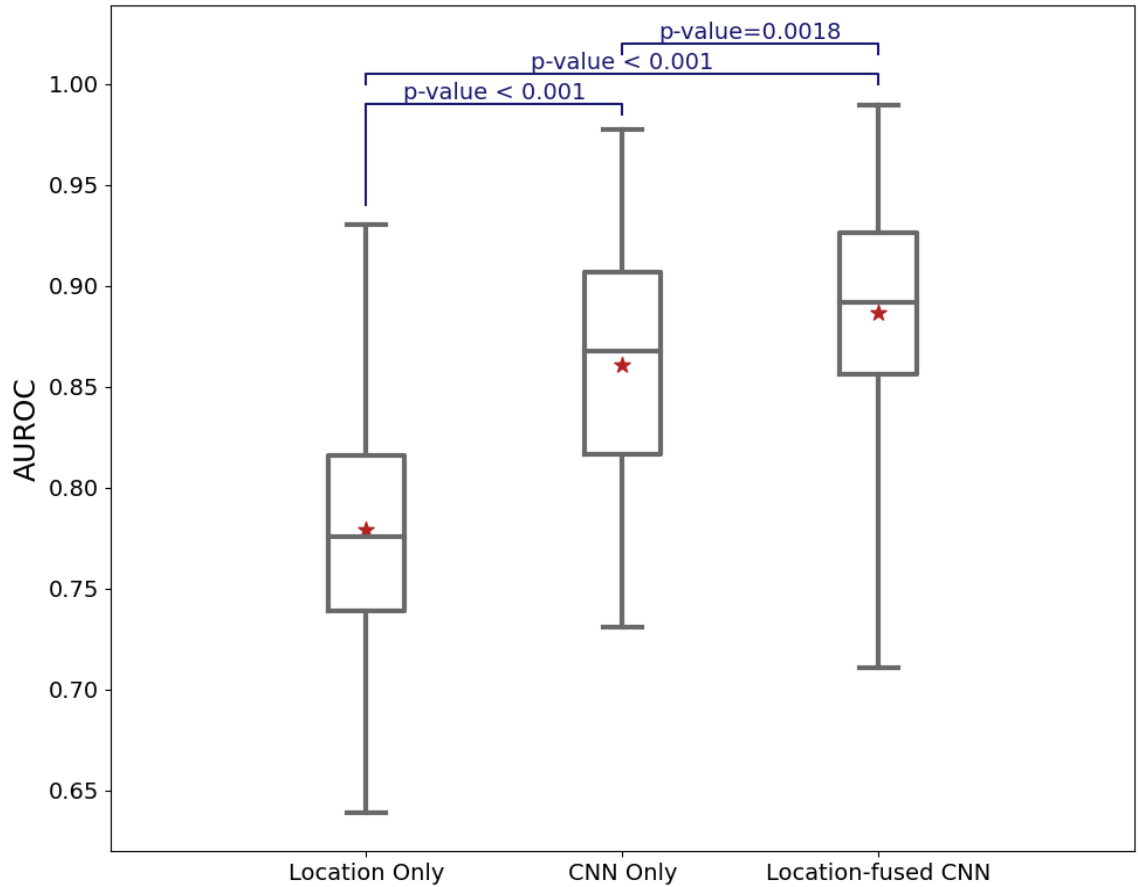

**Figure 3:** AUROC performance of the classification algorithms on test cohorts

# Conclusion

Conventional CNN models lose location information, and in contexts such as pLGG subtype classification where tumor location is informative, the models' performance remains suboptimal. In this paper we proposed an efficient method for reinforcing the tumor location on CNNs' input and achieved statistically significant improvements. The method adds to model's explainability and proves the manually segmented ROIs are not optimal.

## Potential Negative Societal Impact

Blind adoption of Artificial Intelligence, in sensitive contexts such as healthcare, can result in suboptimal planning. Although we measured the randomness of the models and systematically investigated the effect of model initialization and data split, larger datasets, multi-institutional data, and prospective studies are needed to verify generalizability of the models.

## Acknowledgement

This research has been made possible with the financial support of the Canadian Institutes of Health Research (CIHR) (Funding Reference Number: 184015)

# Appendix A

The CNN architecture is visualized in Figure 4, which is generated using the PyTorchViz Python library (https://github.com/szagoruyko/pytorchviz).

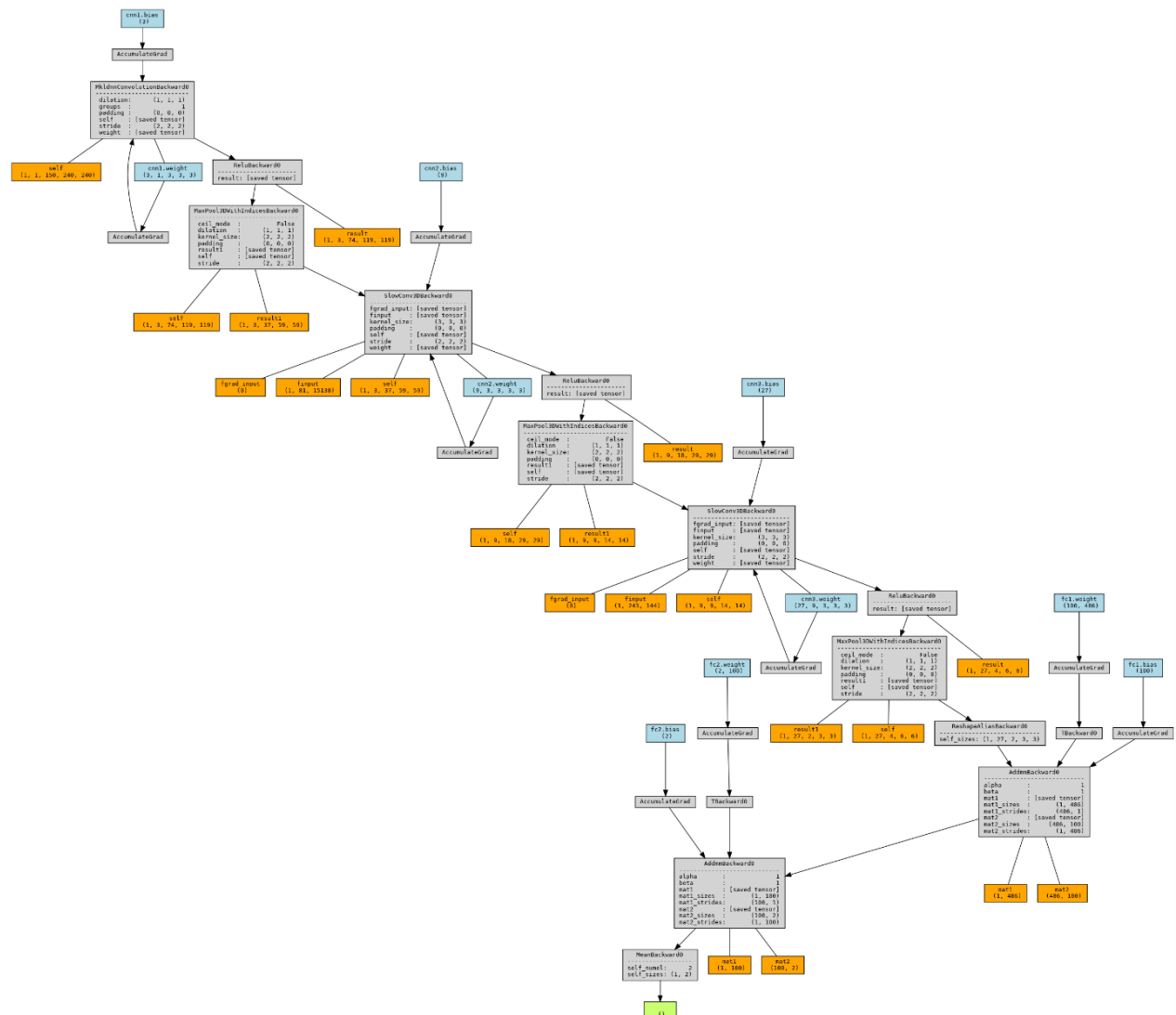

**Figure 4:** Visualization of the CNN architecture and parameters